\def\year{2018}\relax
\documentclass[letterpaper]{article} %
\usepackage{aaai18}  %
\usepackage{times}  %
\usepackage{helvet}  %
\usepackage{courier}  %
\usepackage{url}  %
\usepackage{graphicx}  %
\usepackage{amssymb,amsfonts,amsmath}
\usepackage[mathscr]{euscript}
\usepackage{subcaption}
\usepackage{dsfont}
\usepackage[ruled]{algorithm2e}
\usepackage[textsize=footnotesize,color=green!40]{todonotes}

\usepackage{amssymb,amsmath}
\usepackage{amsthm}
\usepackage{bm}
\def\given{\middle\vert}

\def\indicator{\mathds{1}}

\def\expectation{\mathbb{E}}

\def\prob{P}
\def\defeq{\dot=}
\newcommand{\deriv}[2][]{\frac{\partial#1}{\partial#2}}

\def\state{s}
\def\states{\mathscr{S}}

\def\actions{\mathscr{A}}
\def\option{o}
\def\opt{\option}
\def\Option{O}
\def\Opt{\Option}
\def\options{\mathscr{O}}

\def\astate{z}
\def\Astate{Z}
\newcommand{\R}{\mathbb{R}}

\begin{document}

\title{When Waiting is not an Option : Learning Options with a Deliberation Cost}
\author{Jean Harb$^{*}$, Pierre-Luc Bacon\thanks{\hspace{2mm}Indicates equal contribution.}, Martin Klissarov, Doina Precup\\
Reasoning and Learning Lab, McGill University \\
{\tt \{jharb,pbacon,mklissa,dprecup\}@cs.mcgill.ca}
}

\nocopyright
\maketitle

\begin{abstract}
Recent work has shown that temporally extended actions (options) can be learned fully end-to-end as opposed to being specified in advance. While the problem of \textit{how} to learn options is increasingly well understood, the question of \textit{what} good options should be has remained elusive. We formulate our answer to what \textit{good} options should be in the bounded rationality framework (Simon, 1957) through the notion of \textit{deliberation cost}. We then derive practical gradient-based learning algorithms to implement this objective. Our results in the Arcade Learning Environment (ALE) show increased performance and interpretability.
\end{abstract}

\section{Introduction}

Temporal abstraction has a rich history in AI \cite{Minsky1961,Fikes1972,Kuipers1979,Korf1983,Iba1989,Drescher1991,Dayan1992,Kaelbling1993,Thrun95,Parr1997,Dietterich1998} and has been presented as a useful mechanism for a variety of problems that affect AI systems in may settings, including to: generate shorter plans, speed up planning, improve generalization, yield better exploration, increase robustness against model mis-specification or partial observability. In reinforcement learning, \textit{options} \cite{SuttonPrecupSingh1999} provide a framework to represent, learn and plan with temporally extended actions. Interest in temporal abstraction in reinforcement learning has increased substantially in the last couple of years, due to increasing success in constructing such abstractions automatically from data, e.g. \cite{Bacon2017,Kulkarni2016,Daniel2016,Mankowitz2016,Machado2017}.
However, defining what constitutes a good set of options remains an open problem.

In this paper, we aim to leverage the bounded rationality framework \cite{Simon1957} in order to explain what would make good temporal abstractions for an RL system. A lot of existing reinforcement learning work has focused on Markov Decision Processes, where optimal policies can be obtained under certain assumptions. However, optimality does not take into account possible resource limitations of the agent, which is assumed to have access to a lot of data and computation time. Indeed, options help agents overcome such limitations, by allowing policies to be computed faster \cite{Dietterich1998,Precup2000}. However, from the point of view of absolute optimality, temporal abstractions are not necessary: the optimal policy is achieved by primitive actions. Therefore, it has been difficult to formalize in what precise theoretical sense temporally abstract actions are helpful.

Bounded rationality is a very important framework for understanding rationality in both natural and artificial systems. In this paper, we propose bounded rationality as a lens through which we can describe the desiderata for constructing temporal abstractions, as their goal is mainly to help agents which are restricted in terms of computation time. This perspective helps us to formulate more precisely what objective criteria that should be fulfilled during option construction.  We propose that good options are those which allow an agent to learn and plan \textit{faster}, and provide an optimization objective for learning options based on this idea. We implement the optimization using the option-critic framework \cite{Bacon2017} and illustrate its usefulness with experiments in Atari games.

\section{Preliminaries}

A finite discounted Markov Decision Process $\mathcal{M}$ is a tuple $\mathcal{M} \defeq (\states, \actions, \gamma, r, P)$ where  $\states$ and $\actions$ denote the state and action set respectively, and $\gamma\in [0,1)$ is a discount factor. The reward function $r$ is often assumed to be a deterministic function of the state and actions, but can also map to a distribution, $r : \states \times \actions \rightarrow Dist(\mathbb{R})$ (a perspective which we use in our formulation). The transition matrix $P: \states \times \actions \to Dist(\states)$ is a conditional distribution over next states given that an action $a \in \actions$ is taken under a certain state $s \in \states$. The interaction of a randomized stationary policy $\pi: \states \to Dist(\actions )$ or a deterministic policy $\pi: \states \to \actions$ with an MDP $\mathcal{M}$ induces a Markov process over states, actions and rewards over which is defined the expected discounted return $V_\pi(s) \defeq \expectation_\pi\left[ \sum_{t=0} \gamma^t r(S_t, A_t) \given S_0 = s\right]$. The value function $V_\pi$ of a policy $\pi$ satisfies the Bellman equations :
\begin{align*}
    V_\pi(s) = \sum_{a} \pi\left(a \given s\right)\left( r(s, a) + \gamma \sum_{s'} \prob\left(s' \given s, a\right) V_\pi(s')\right)
\end{align*}
In the control problem, we are interested in finding an \textit{optimal policy} for a given MDP. A policy $\pi^\star$ is said to be optimal if $V_{\pi^\star}(s) \defeq \max_{\pi} V_\pi(s)$ for all $s$.

An important class of control methods in reinforcement learning is based on the actor-critic architecture \cite{Sutton1984}. In the same way that function approximation can be used for value functions, policies can also be approximated within a parameterized family which is searched over. In the policy gradient theorem, \cite{Sutton1999} shows that the gradient of the expected discounted return  with respect to the parameters of a policy is of the form $\expectation_{\alpha,\pi_\theta}\left[\sum_{a} \deriv[\pi_\theta\left(a \given s\right)]{\theta}Q_{\pi_\theta}(s, a)\right]$, where $\alpha$ is an initial state distribution. A locally optimal policy can then be found by stochastic gradient ascent over the policy parameters while simultaneously learning the action-value function $Q_{\pi_\theta}(s,a)$ (usually by TD).

\subsection{Options}

Options \cite{SuttonPrecupSingh1999} provide a framework for representing, planning and learning with temporally abstraction actions. The option frameworks assumes the existence of a base MDP on which are overlaid temporally abstract actions called \textit{options}. An option is defined as a triple $(\mathcal{I}_\opt, \pi_\opt, \beta_\opt)$ where $\mathcal{I} \subseteq \states$ is an initiation set, $\pi_\opt: \states \to Dist(\actions)$ is the policy of an option  (which can also be deterministic) and $\beta_\opt : \states \to [0, 1]$ is a termination condition. In the call-and-return execution model, a policy over options $\mu: \states \rightarrow  Dist(\options)$ (deterministic if wanted) chooses an option among those which can be initiated in a given state and executes the policy of that option until termination. Once the chosen option has terminated, the policy over options chooses a new option and the process is repeated until the end of the episode.

The combination of a set of options and base MDP leads to a semi-Markov decision process (SMDP) \cite{Howard1963,Puterman1994} in which the transition time between two decision points is a random variable. When considering the induced process only at the level of state-option pairs, usual dynamic programming results can be reused after a transformation to an equivalent MDP \cite{Puterman1994}. To see this, we need to define two kinds of \textit{models} for every option : a reward model $b_\opt : \states \to \R$ and a transition model $F_\opt : \states \times \states \to \R$. If an option does not depend on the history since initiation, we can write its models either in closed form or as the solution to Bellman-like equations \cite{SuttonPrecupSingh1999}. The expected discounted return associated with a set of options $\options$ and a policy over them is the solution $Q_\theta$ to a set of Bellman equations :
\begin{align*}
    Q_\theta(s, \opt) &= b_\theta(s,\opt) + \sum_{s'} F_\theta(s', s, \opt) V_\theta(s')\\
    &\defeq \expectation_\theta\left[\sum_{t=0} \gamma^t r(S_t, A_t) \given S_0 = s, \Opt_0 = \opt \right]
\end{align*}
where $\theta$ is a concatenation of the policy over options $\mu$, options policies and termination conditions.

\subsection{Intra-Option Bellman Equations}
In the case of Markov options, there exists another form for the Bellman equations, called the intra-option Bellman equations \cite{SuttonPrecupSingh1999}, which are key for deriving gradient-based algorithms for learning options. %

Let $Z_t \defeq (S_t, \Opt_t)$ be a random variable over state-option tuples. We call the space of state-option pairs the {\em augmented state space}. This augmentation is sufficient to provide the Markov property, which would otherwise be lost when considering the process at the \textit{flat} level of state-action pairs \cite{SuttonPrecupSingh1999}.  The transition matrix of the Markov process over the augmented state space \cite{Bacon2017} is given by :
\begin{align*}
&\widetilde{\prob}_\theta\left(\astate' \given \astate, a\right) = \\
&\prob\left(s' \given s, a\right)\left( (1 - \beta_\theta(s', \opt))\indicator_{\opt'=\opt} + \beta_\theta(s', \opt) \mu_\theta\left(\opt' \given s'\right) \right) \enspace .
\end{align*}
Using this chain structure, we can define the MDP $\widetilde{\mathcal{M}} \defeq (\widetilde{P}_\theta, \widetilde{r}, \gamma)$ whose associated value function  $\widetilde{V}_\theta : (\states \times \options) \to \R$ is:
\begin{align}
&\widetilde{V}_\theta(\astate) = \expectation_\theta\left[\sum_{t=0}^\infty \gamma^t \widetilde{r}(\Astate_t, A_t,\Astate_{t+1}) \given \Astate_0 = \astate\right]\notag\\
&=\sum_{a,\astate'} \pi_\theta\left(a \given \astate\right) \widetilde{\prob}_\theta\left( \astate' \given \astate, a\right) \left( \widetilde{r}(\astate, a, \astate') + \gamma \widetilde{V}_\theta(\astate')\right)  \enspace . \label{eq:augbell}
\end{align}
Since the rewards come from the base (primitive) MDP, we can simply write $\widetilde{r}(z, a, z') = r(s,a)$ and because $\sum_{\astate'}\widetilde{\prob}_\theta\left( \astate' \given \astate, a\right) = 1$, we get:
\begin{align*}
\sum_{\astate'}\widetilde{\prob}_\theta\left( \astate' \given \astate, a\right) \widetilde{r}(\astate, a, \astate') &= r(s, a)\enspace .
\end{align*}
Hence, when taking the expectation in \eqref{eq:augbell} over the next values, we obtain :
\begin{align}
&\widetilde{V}_\theta(\astate) \defeq Q_\theta(s, \opt) =\sum_{a} \pi_\theta\left(a \given s, \opt\right)\bigg(r(s, a) + \notag \\
&\gamma \sum_{s'} \prob\left(s' \given s, a\right) \Big[Q_\theta(s', \opt) - \beta_\theta(s', \opt) A_\theta(s', \opt)\Big]\bigg)\enspace , \label{eq:intra}
\end{align}
where $A_\theta(s,\opt) \defeq Q_\theta(s,\opt)  - V_\theta(s)$ is the advantage function \cite{Baird1993}. The equations in \eqref{eq:intra} correspond exactly to the intra-option Bellman equations \cite{SuttonPrecupSingh1999}. However, we chose to present them under an alternate -- but more convenient -- form highlighting a connection to the advantage function:
\begin{align*}
U(s', \opt) &\defeq (1 - \beta_\theta(s',\opt))Q_\theta(s', \opt) + \beta_\theta(s') V_\theta(s')  \\
&= Q_\theta(s', \opt) - \beta_\theta(s', \opt) A_\theta(s', \opt) \enspace ,
\end{align*}
where $U(s', \opt)$  represents the \textit{utility} of continuing with the same option or switching to a better one.

\subsection{Optimization}

The option-critic architecture \cite{Bacon2017} is a gradient-based actor-critic architecture for learning options end-to-end. As in actor-critic methods, the idea is to parametrize the options policies and termination conditions and learn their parameters jointly by stochastic gradient ascent on the expected discounted return. \cite{Bacon2017} provided the form of the gradients for both the option policies and termination \textit{functions} under the assumption that options are available everywhere. In the following, we further assume that the parameter vector $\theta = [\theta_\mu ; \theta_\pi ; \theta_\beta]$ is partitioned into disjoint sets of parameters for the policy over option, the option policies and the termination functions.

In the gradient theorem for options policies \cite{Bacon2017}, the result maintains the same form as that of original policy gradient theorem for MDP \cite{Sutton1999} but over the augmented state space. If $J_\alpha(\theta)$ is the expected discount return for the set of options and the policy over them, then the gradient of the option policies (whose parameters are independent from the terminations) is :
\begin{align*}
    \deriv[J_\alpha(\theta)]{\theta_\pi} =  \gamma \expectation_{\alpha, \theta}\left[\sum_{a} \deriv[\pi_\theta\left(a \given \astate\right)]{\theta_\pi} \widetilde{Q}_\theta(\astate, a)\right] \enspace ,
\end{align*}
where $\alpha$ is an initial state distribution over state and options.

To obtain the gradient for the termination functions, let's first take the derivative of the intra-option Bellman equations:
\begin{align}
    &\deriv[Q_\theta(s,\opt)]{\theta_\beta}= \gamma\sum_{a} \pi_\theta\left(a \given s, \opt\right)  \sum_{s'} \prob\left(s' \given s, a\right) \Bigg[  \notag \\
    &- \deriv[\beta_\theta(s', \opt)]{\theta_\beta} A_\theta(s', \opt) +\deriv[Q_\theta(s,\opt)]{\theta_\beta} - \beta_\theta(s',\opt) \deriv[A_\theta(s',\opt)]{\theta_\beta} \Bigg] \label{eq:derivbellman}
\end{align}
By noticing the similarity between \eqref{eq:derivbellman} and \eqref{eq:augbell}, we can easily solve for the recursive form of the derivative. Indeed, it suffices to see that $ -\deriv[\beta_\theta(s', \opt)]{\theta_\beta} A_\theta(s', \opt)$ plays the role of the ``reward'' term in the usual Bellman equations (see \cite{Bacon2017} for a detailed proof) and conclude that:
\begin{align}
    \deriv[J_\alpha(\theta)]{\theta_\beta} = \gamma \expectation_{\alpha, \theta}\left[ -\deriv[\beta_\theta(s', \opt)]{\theta_\beta} A_\theta(s',\opt) \right] \label{eq:termgrad}
\end{align}
Hence the termination gradient shows that if an option is advantageous, the probability of termination should be lowered, making that option longer. Conversely, if the value of an option is less than what could be achieved through a different choice of option at a given state, the termination gradient will make it more likely to terminate at this state.
The termination gradient has the same structure as the interruption operator \cite{Mann2014} in the \textit{interruption execution} model \cite{SuttonPrecupSingh1999}. Rather than executing the policy of an option irrevocably until termination, interruption execution consists in choosing a new option whenever $Q_\theta(s,\opt) < V_\theta(s)$. Moving the the value function $V_\theta$ to the left hand side, interruption execution can also be understood in terms of the advantage function: $Q_\theta(s,\opt) < V_\theta(s) \Leftrightarrow A_\theta(s, \opt) < 0$. As for the termination gradient, interruption execution leads to the termination of an option whenever there is no advantage (negative advantage) in maintaining it. Interestingly, \cite{Mann2014} also considered adding a scalar \textit{regularizer} to the advantage function to favor longer options. From the more general perspective of bounded rationality, we also recover this regularizer but within a larger family which follows from the notion of deliberation cost.

\section{Deliberation Cost Model}

From a representation learning perspective, good options ought to allow an agent to learn and plan \textit{faster} \cite{Minsky1961}. Due to their temporal structure, options offer a mechanism through which an agent can make better use of its limited computational resources and act faster. Once an option has been chosen, we assume that the computational cost of executing that option is negligible or constant until termination. After deliberating on the choice of option, an agent can \textit{relax} thanks to the fast -- but perhaps imperfect -- knowledge compiled within the its policy.

This perspective on options is similar to \textit{fast and frugal} heuristics \cite{Gigerenzer2001} which form a decision repertoire for efficient decision making under limited resource. Our assumption on the cost structure is also consistent with models of the prefrontal areas \cite{Botvinick2009,Solway2014} presenting  decision making over options as a slower model-based planning process as opposed to fast and habitual learning taking place within an option. When planning with options (in computers), there is also a cost for deciding which option to choose next by making predictions based on their models. For example, options models could be given by deep networks, necessitating back-and-forth to the GPU, or using a simulator with costly explicit rollouts \cite{Guo2014,Mann2015}.

Bounded rationality can also be useful to understand how efficient communication can take place between two agents over a limited channel \cite{Neyman1985}. Options offer a mechanism for communicating intents to and from an agent \cite{Branavan2012,Andreas2017} more efficiently, by compressing the information into a simpler form, sending only the identifier of the options and not the details themselves. Having longer options is a way to provide better interpretability and simplifies communication and understanding by compressing information.

\begin{figure}
    \def\svgwidth{\columnwidth}
    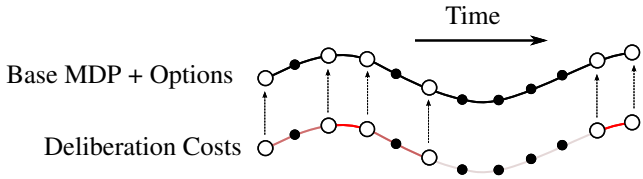
    \caption{A deliberation cost is incurred upon switching to a new option and is subtracted from the reward of the base MDP. Open circles represent SMDP decision points while filled circles are primitive steps within an option. The cost rate for each option is represented by the intensity of the subtrajectory.}
    \label{fig:delibcost}
\end{figure}

Consider the cost  model (fig. \ref{fig:delibcost}) in which executing an option within an option is free but switching to an option upon arriving in a new state incurs a cost $\eta$ .
To build some intuition, let's further assume that the termination function of an option is constant over all states. If $\kappa$ is the continuation probability of that option, its expected discounted duration is $d=\frac{1}{1 - \gamma \kappa}$. When a fixed cost $\eta$ is incurred upon termination, the average cost per step for that option is then $\eta/d = (1 - \gamma \kappa)\eta$. Hence, as the probability of continuation increases and options get longer, the cost rate decreases. Conversely, if an option only terminates after one step -- a primitive option -- $\kappa$ is $0$ and the cost rate is $\eta$. The fact that longer options lead to a better amortization of the deliberation cost is key to understanding their benefit in comparison to using only primitive actions.

\subsection{Formulation}

In addition to the value function $\widetilde{V}_\theta(z) \defeq Q_\theta(s, \opt)$ for the base MDP and options over them, we define an immediate cost function $\widetilde{c}(z, a, z') \defeq c(s, \opt, a, s', \opt')$ and a corresponding deliberation cost function $\widetilde{D}_\theta(z) \defeq D_\theta(s, \opt)$. The expected sum of discounted costs associated with a set of options and the policy over them is given by the function $\widetilde{D}_\theta : (\states \times \options) \to \R$:
\begin{align*}
    \widetilde{D}_\theta(\astate) = \expectation_\theta\left[\sum_{t=0}^\infty  \gamma^t \widetilde{c}(\Astate_t, A_t, \Astate_{t+1}) \given \Astate_0 = \astate\right]
\end{align*}

We first formulate our goal of maximizing the expected return while keeping the deliberation cost low as a constrained optimization problem:
\begin{align*}
&\max_{\theta}  \sum_{s,\opt} \alpha(s,\opt) Q_\theta(s,\opt) \\
&\text{subject to :} \sum_{s,\opt} \alpha(s,\opt) D_\theta(s,\opt) \leq k\enspace,
\end{align*}
where $\alpha$ is an initial state distribution over state-option pairs. But in general, solving a problem of this form \cite{Altman1999} requires a Linear Programming (LP) formulation which is both expensive to solve and incompatible with the model-free learning methods adopted in this work. Instead, we consider the unconstrained optimization problem arising from the Lagrangian formulation \cite{Sennott1991,Altman1999}:
\begin{align}
    &\max_\theta  J_\alpha(\theta)\enspace , \notag \\
    &\text{where } J_\alpha(\theta) \defeq \sum_{s,\opt} \alpha(s,\opt) \left(Q_\theta(s, \opt) - \eta D_\theta(s,\opt)\right), \label{eq:optim}
\end{align}
and $\eta \in \R$ is a regularization coefficient. While \eqref{eq:optim} shows the option-value function and the deliberation cost function as separate entities, they can in fact bee seen as a single MDP whose reward function is the difference of the base MDP reward and the cost function:
\begin{align*}
 J_\alpha(\theta) &= \expectation_{\alpha,\theta}\left[\sum_{t=0}^\infty  \gamma^t \widetilde{r}\left(\Astate_t, A_t, \Astate_{t+1}\right) - \gamma^t \eta \widetilde{c}\left(\Astate_t, A_t, \Astate_{t+1}\right) \right].
\end{align*}
Therefore, there is a set of Bellman equations which the value function $\widetilde{V}_\theta^c(\astate)\defeq Q_\theta^c(s,\opt)$ over the transformed reward function satisfies:
\begin{align}
    \widetilde{V}_\theta^c(\astate) \defeq \sum_{a,\astate'} &\pi_\theta\left(a \given s, \astate\right)\widetilde{\prob}_\theta\left(\astate' \given \astate, a, \right) \Big( \notag \\
    &\widetilde{r}(\state, a, \astate') - \eta \widetilde{c}(\astate, a, \astate') + \gamma \widetilde{V}_\theta^c(\astate')   \Big) \enspace . \label{eq:augtrans}
\end{align}
Similarly, there exist Bellman optimality equations in the sense of \cite{SuttonPrecupSingh1999} for the parameters of the policy over options $\theta_\mu$:
\begin{align}
    Q_{\options}^\star(s, \opt) \defeq \max_{\theta_\mu \in \Pi(\options)} \left(Q_{\theta_\mu}(s, \opt) - \eta D_{\theta_\mu}(s,\opt)\right)  \enspace . \label{eq:opt}
\end{align}
where the notation $Q_{\theta_\mu}$ here indicates that the parameters for the options are kept fixed and only $\theta_\mu$ is allowed to change. A policy over option $\mu^\star$ is $\eta$-optimal with respect to a set of options if it reaches the maximum in \eqref{eq:opt} for a given $\eta$. Clearly, when  $\eta=0$, the corresponding $\mu^\star$ is also optimal in the base MDP and there is no loss of optimality in this regard.

\subsection{Switching Cost and its Interpretation as a Margin}

One way to favor long options is by a cost function which penalizes for frequent options switches. In the same way that the MDP formulation allows for randomized reward functions \cite{Puterman1994}, we can also capture the random event of switching through the immediate cost function $c$. Since $\beta_\theta(s', \opt)$ is the mean of a Bernoulli random variable over the two possible outcomes, switching  or continuing ($1$ or $0$), the cost function corresponding to the switching event is $c_\theta(s', \opt) = \gamma \beta_\theta(s', \opt)$ (where $\gamma$ was added for mathematical convenience).

When expanding the value function  over the transformed reward \eqref{eq:augtrans} for this choice of $c_\theta(s', \opt)$, we get:
\begin{align}
    Q_\theta^c(s,\opt) = &\sum_{a} \pi_\theta\left(a \given s, \opt\right) \bigg( r(s,a) + \gamma \sum_{s'} \prob\left(s' \given s, a\right) \Big[ \notag \\
    & Q_\theta^c(s', \opt)  - \beta_\theta(s', \opt)\left( A_\theta^c(s, \opt) + \eta\right)\Big] \bigg) . \label{eq:bellman-cost}
\end{align}
with $\eta$ appearing along with the advantage function : a term which would otherwise be absent from the intra-option Bellman equations over the base MDP \eqref{eq:intra}. Therefore, adding the switching cost function to the base MDP reward contributes a scalar \textit{margin} $\eta$ to the advantage function $A_\theta^c$ over the transformed reward. When learning termination functions in option-critic, the termination gradient for the unconstrained problem \eqref{eq:optim} is then of the form:
\begin{align}
    \deriv[J_\alpha(\theta)]{\theta_\beta} = \gamma \expectation_{\alpha,\theta}\left[- \deriv[\beta_\theta(S_{t+1}, \Opt_t)]{\theta_\beta} \left(A_\theta^c(S_{t+1}, \Opt_t)+ \eta\right)\right] \enspace  . \label{eq:gradtrans}
\end{align}
Hence, $\eta$ sets a \textit{margin} or a \textit{baseline} for how good an option ought to be : a correction which might be due to approximation error or to reflect some form of uncertainty in the value estimates. By increasing its value, we can reduce the gap in the advantage function, tilting the balance in favor of maintaining an option rather than terminating it.

\subsection{Computational Horizon}

Due to the generality of our formulation, the discount factor of the deliberation cost function can be different from that of the value function over the base MDP reward. The unconstrained formulation of \eqref{eq:optim} then becomes a function of two discount factors: $\gamma$ for base MDP and $\lambda$ for the deliberation cost function:
\begin{align*}
    J_\alpha^{\gamma,\lambda}(\theta) &= \sum_{s, \opt}\alpha(s,\opt)\left(Q_\theta^\gamma(s, \opt) - \eta D_\theta^\lambda(s,\opt)\right)
\end{align*}
Since the derivative of the deliberation cost function with respect to the termination parameters is:
\begin{align*}
    &\deriv[D_\theta^\lambda(s,\opt)]{\theta_\beta}= \deriv[]{\theta_\beta} \sum_a \pi_\theta\left(a \given s, \opt\right) \sum_{s'}\prob\left(s' \given s, a\right)\Big( c_\theta(s', \opt) +\\
    & \lambda \big[(1 - \beta_\theta(s',\opt)D_\theta^\lambda(s',\opt) + \beta_\theta(s',\opt)\sum_{\opt'}\mu_\theta\left(\opt'\given s'\right) D_\theta^\lambda(s', \opt')\big]\Big) ,
\end{align*}
setting $\lambda=0$ when the cost function is $c_\theta(s',\opt) \defeq \gamma \beta_\theta(s',\opt)$ leaves only one term : $\deriv[\beta_\theta(s',\opt)]{\theta_\beta} \eta$. Hence, by linearity with \eqref{eq:termgrad}, the derivative over the mixed objective is:
\begin{align}
    \deriv[J_\alpha^{\gamma,\lambda=0}(\theta)]{\theta_\beta}  &= \gamma \expectation_{\alpha, \theta}\bigg[ -\deriv[\beta_\theta(s',\opt)]{\theta_\beta}\Big(A_\theta(s',\opt) + \eta\Big)\bigg] \enspace  . \label{eq:gradtranslambdazero}
\end{align}
While similar to \eqref{eq:gradtrans} in the sense that the margin $\eta$ also enters the advantage function, \eqref{eq:gradtranslambdazero} differs fundamentally in the fact that it depends on $A_\theta$ and not $A_\theta^c$, the advantage function over the transformed reward. We can also see that when $\gamma = \lambda$, we recover the same form for the derivative of the expected return in the transformed MDP from \eqref{eq:gradtrans}:
\begin{align*}
    \deriv[J_\alpha^{\gamma=\lambda}(\theta)]{\theta_\beta}  &= \gamma \expectation_{\alpha, \theta}\bigg[ -\deriv[\beta_\theta(s',\opt)]{\theta_\beta}\left(A_\theta^c(s',\opt) + \eta\right)\bigg] \\
    &= \deriv[J_\alpha(\theta)]{\theta_\beta} \enspace .
\end{align*}
The discount factor $\lambda$ for the deliberation cost function provides a mechanism for truncating the sum of costs. Therefore, it plays a distinct role from the regularization coefficient $\eta$ which merely scales the deliberation cost function but does not affect the \textit{computational horizon}. As opposed to the random horizon set by the discount factor $\gamma$ in the environment, $\lambda$ pertains to the \textit{internal} environment of agent about  the cost of its own \textit{cognitive} or computational processes. It is a parameter about an introspective process of self-prediction of how likely a sequence of internal costs will be accumulated as a result of deliberating about courses of action in the \textit{outside} environment. In accordance with more general results on discounting \cite{Petrik2008,Jiang2015}, $\lambda$ should be aligned with the representational capacity of the system since $\lambda \to 1$ involves an increasingly more difficult prediction problem.

In that sense, $\lambda=0$ indicates that only the immediate computational cost should be considered when learning options that also maximize for the reward. When learning termination functions, the resulting shallow evaluation under small values of $\lambda$ might not take into account the possibility that the overall expected cost could be lowered in exchange of a less favorable immediate cost : it lacks foresight.  Despite the fact that the full effect of a change in the options or the policy over them cannot be captured with $\lambda=0$, the corresponding gradient \eqref{eq:gradtrans} is still useful when $\eta > 0$. It leads to both the regularization strategy proposed in \cite{Bacon2017} for gradient-based learning and \cite{Mann2014} in the dynamic programming case. Furthermore, since \eqref{eq:gradtrans} does not depend on $A^c_\theta$, values can be learned only for the original reward function and does not require mixed or separate estimates.

\section{Experiments}

Previous results \cite{Bacon2017} in the Arcade Learning Environment \cite{Bellemare2013} have shown that while learning options end-to-end is possible, frequent terminations can become an issue unless regularization is used. Hence, we chose to apply the idea of deliberation cost in combination with a novel option-critic implementation based on the Advantage Asynchronous Actor-Critic (A3C) architecture of \cite{mnih2016asynchronous}.  More specifically, our experiments are meant to assess  : the interpretability of the resulting options, whether degeneracies (frequent terminations) to single-step options can be controlled, and if the deliberation cost can provide an inductive bias for learning faster.

\subsection{Asynchronous Advantage Option-Critic (A2OC)}

\begin{algorithm}[h]
\caption{Asynchronous Advantage Option-Critic}
\label{alg:oc-intraq2}
\DontPrintSemicolon
Initialize global counter $T \leftarrow 1$\;
Initialize thread counter $t \leftarrow 1$\;
$c \leftarrow 0$\;
\Repeat{$T > T_{\max}$}{
$t_{start} = t$\;
$s_t \leftarrow s_0$ \;
Reset gradients: $dw \leftarrow 0$, $d\theta_\beta \leftarrow 0$ and $d\theta_\pi \leftarrow 0$\;
Choose $\option_t$ with an $\epsilon\text{-soft}$ policy over options $\mu(s_t)$\;

\Repeat{episode ends or $t-t_{start} == t_{max}$ or ($t-t_{start} > t_{min}$ and  $\option_{t}$ terminated)} {
Choose $a_t$ according to $\pi_{\theta}\left( \cdot \given s_t \right)$\;
Take action $a_t$ in $s_t$, observe $r_t, s_{t+1}$\;
$\widetilde{r_t} \leftarrow r_t + c_t$\;

\uIf{the current option $\opt_t$ terminates in $s_{t+1}$}{
    choose new $\opt_{t+1}$ with $\epsilon\text{-soft}(\mu(s_{t+1}))$\;
$c \leftarrow \eta$}
\Else{$c \leftarrow 0$}
$t \leftarrow t + 1$\;
$T \leftarrow T + 1$\;
}
$G = V_\theta(s_t)$\;
\For{$k \in {t-1, ... , t_{start}}$}{
$G \leftarrow \widetilde{r}_k + \gamma G$\;
Accumulate thread specific gradients:\;
$d w \leftarrow dw - \alpha_{w} \deriv[(G - Q_\theta(s_k, \option_k))^2]{w}$\;
$d\theta_\pi \leftarrow d\theta_\pi + \alpha_{\theta_\pi} \deriv[\log \pi_{\theta}\left( a_k \given s_k\right)]{\theta_\pi} (G - Q_\theta(s_k, \option_k))$\;
$d\theta_\beta \leftarrow d\theta_\beta - \alpha_{\theta_\beta} \deriv[\beta_{\theta}(s_k)]{\theta_\beta} \left( Q_\theta(s_k, \option_k) - V_\theta(s_k) + \eta \right)\;$\;
}
Update global parameters with thread gradients\;
}
\end{algorithm}

The option-critic architecture \cite{Bacon2017} had introduced a deep RL version of the algorithm, which allowed one to learn options in an end-to-end fashion, directly from pixels. However, it was built on top of the DQN algorithm \cite{mnih2015human}, which is an off-line algorithm using samples from an experience replay buffer. Option-critic, on the other hand, is an on-line algorithm which uses every new sampled transition for its updates. Using on-line samples has been known to cause issues when training deep networks.

Recently, the asynchronous advantage actor-critic (A3C) algorithm \cite{mnih2016asynchronous} addressed this issue and lead to to stable on-line learning by running multiple parallel agents. The parallel agents allows the deep networks to see samples from very different states, which greatly stabilizes learning. This algorithm is also much more consistent with the spirit of option-critic, as they both use on-line policy gradients to train. We introduce the \footnote{The source code is available at \url{https://github.com/jeanharb/a2oc_delib}}{asynchronous advantage option-critic (A2OC)}, an algorithm (alg. \ref{alg:oc-intraq2}) that learns options in a similar way to A3C but within the option-critic architecture.

The architecture used for A2OC was kept as consistent with A3C as possible. We use a convolutional neural network of the same size, which outputs a feature vector that is shared among 3 heads as in \cite{Bacon2017}: the option policies, the termination functions and the Q-value networks. The option policies are linear softmax functions, the termination functions use sigmoid activation functions to represent probabilities of terminating and the Q-values are simply linear layers. During training, all gradients are summed together, and updating is performed in a single thread instance. A3C only needs to learn a value function for its policy, as opposed to Q-values for every action. Similarly, A2OC gets away with the action dimension through sampling \cite{Bacon2017} but needs to maintain state-option value because of the underlying augmented state space.

As for the hyperparameters, we use an $\epsilon$-greedy policy over options, with $\epsilon=0.1$. The preprocessing are the same as the A3C, with RGB pixels scaled to $84\times 84$ grayscale images. The agent repeats actions for 4 consecutive moves and receives stacks of 4 frames as inputs. We used entropy regularization of $0.01$, which pushes option policies not to collapse to deterministic policies. A learning rate of $0.0007$ was used in all experiments. We usually trained the agent with $16$ parallel threads.

\subsection{Empirical Effects of Deliberation Cost}

\begin{figure*}[h]
\centering
\begin{subfigure}[t]{.32\textwidth}
  \centering
  \includegraphics[width=0.8\linewidth]{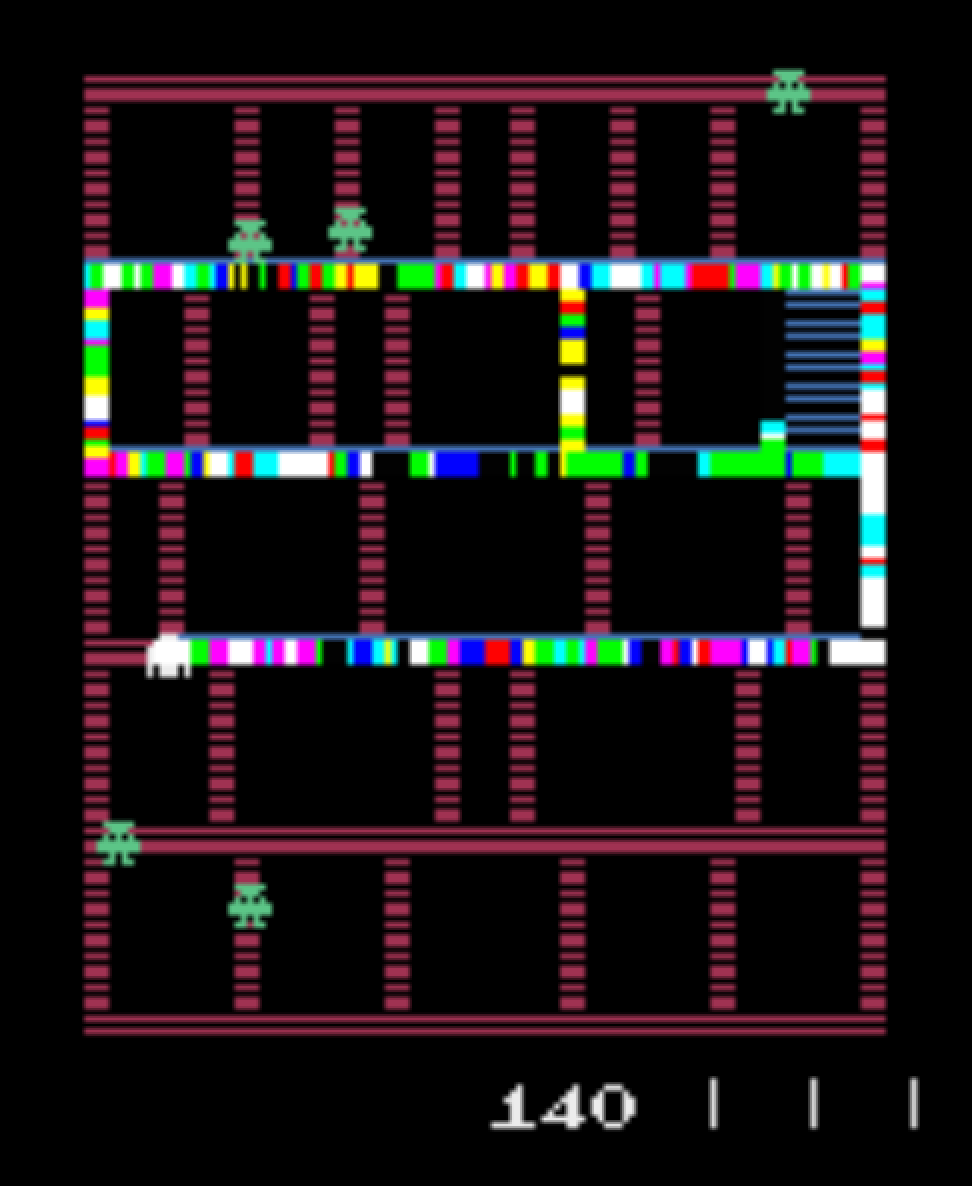}
  \caption{Without a deliberation cost, options terminate instantly and are used in any scenario without specialization.}
  \label{fig:sub3}
\end{subfigure}%
\hfill
\begin{subfigure}[t]{.32\textwidth}
  \centering
  \includegraphics[width=0.8\linewidth]{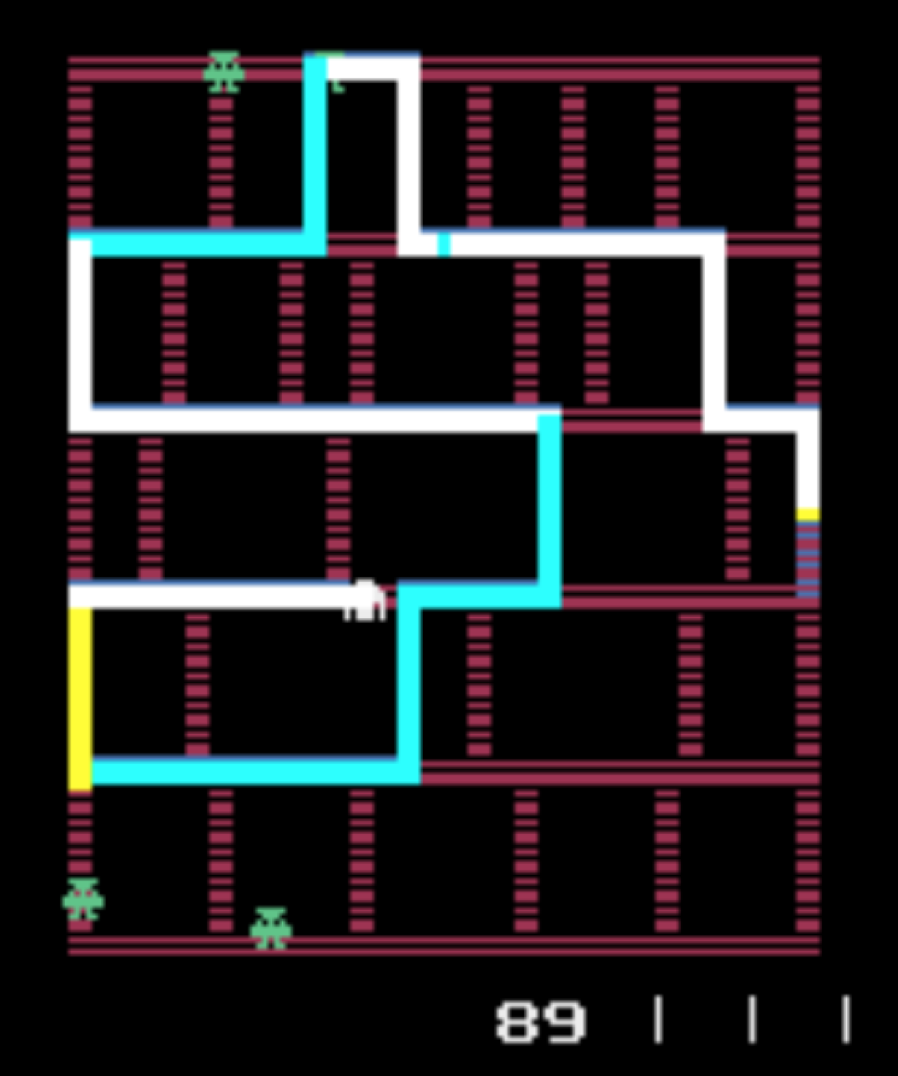}
  \caption{Options are used for extended periods and in specific scenarios through a trajectory, when using a deliberation cost.}
  \label{fig:sub2}
\end{subfigure}%
\hfill
\begin{subfigure}[t]{.32\textwidth}
  \centering
  \includegraphics[width=0.8\linewidth]{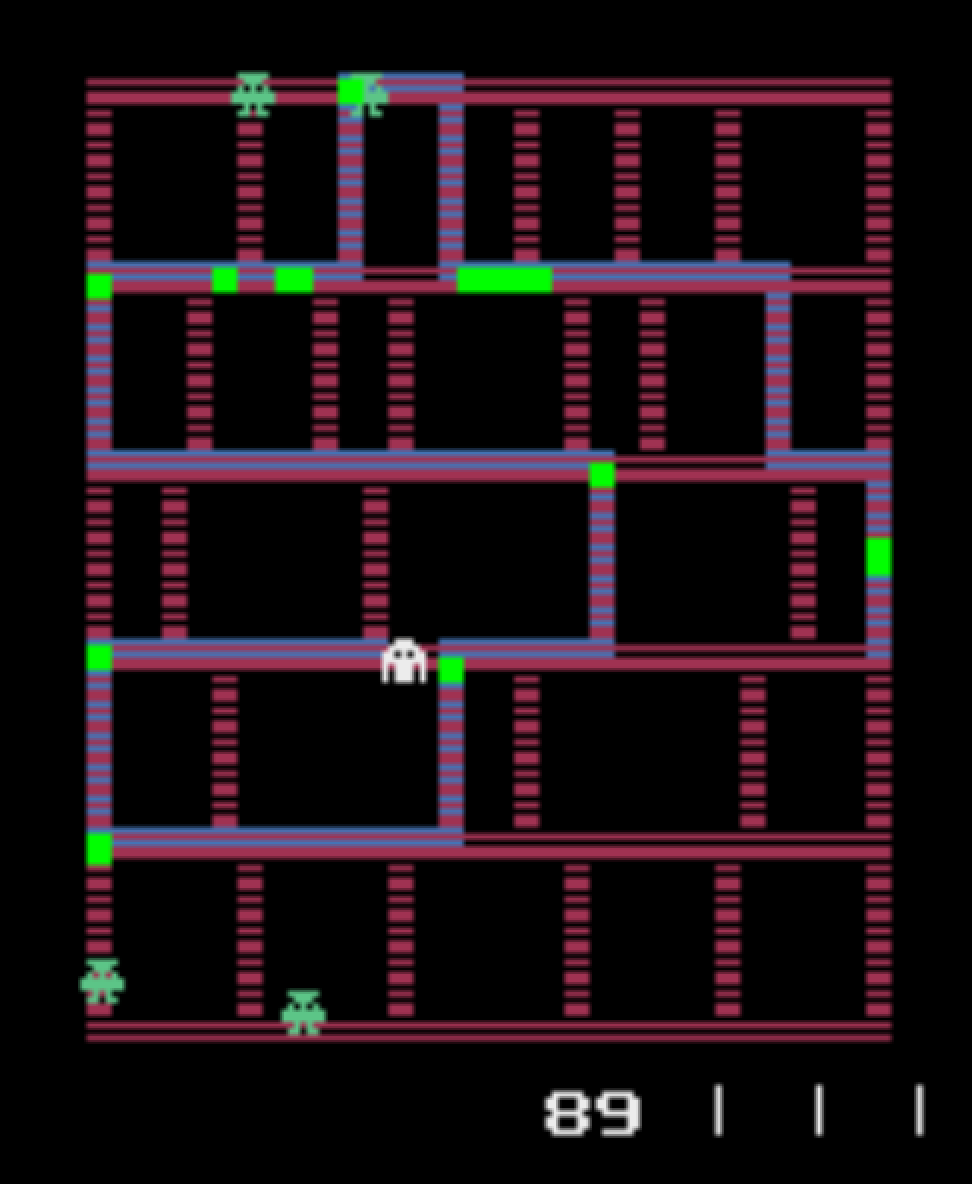}
  \caption{Termination is sparse when using the deliberation cost. The agent terminates options at intersections requiring high level decisions.}
  \label{fig:sub1}
\end{subfigure}%
\caption{We show the effects of using deliberation costs on both the option termination and policies. In figures (a) and (b), every color in the agent trajectory represents a different option being executed. This environment is the game Amidar, of the Atari 2600 suite.}
\label{fig:delibfigs}
\end{figure*}

We use Amidar, a game of the Atari 2600 suite, as an environment that allows us to analyze the option policies and terminations qualitatively. The game is grid-like and the task is to cover as much ground as possible, without running into enemies.

Without a deliberation cost, the options eventually learn to terminate at every step, as seen in figure \ref{fig:sub3}, where every color in the figure represents a different option being executed at the time the agent was in the location.  In contrast, figure \ref{fig:sub2} shows the effect of training an agent with a deliberation cost, which persists with an option over a long period of time. The temporally extended structure of the options shown by color does not result from simply terminating and re-picking the same option at every step but truly represents a contiguous segment of the trajectory where that option was maintained in a call-and-return fashion. Only at certain intersections the options terminate, allowing the agent to select an option which will lead it in a different direction. As opposed to the agent which was trained without a deliberation cost (fig. \ref{fig:sub3}), figure \ref{fig:sub3} shows that the options learned with the regularizer were specialized and only selected in specific scenarios. Figure \ref{fig:sub1} shows us where the agent terminated options on its trajectory. The options are clearly terminating at intersections, which represent key decision points.
\begin{table}[h]
\centering
\resizebox{\columnwidth}{!}{
\begin{tabular}{ |l||c|c|c|c| }
\hline
Algorithm & Amidar & Asterix & Breakout & Hero \\
 \hline
\cite{mnih2015human} & $739.5$ & $6012.0$ & $401.0$ & $19950.0$ \\
\cite{mnih2016asynchronous} & $283.9$ & $6723.0$ & $\boxed{551.6}$ & $\boxed{28765.8}$\\
No deliberation cost & $512.0$ & $1950.0$ & $395.0$ & $2625.0$  \\
$\lambda=0, \eta=0.010$ & $535.0$ & $4700.0$ & $421.0$ & $19805.0$  \\
$\lambda=0, \eta=0.020$ & $\boxed{880.0}$ & $5400.0$ & $\mathbf{430.0}$ & $\mathbf{20100.0}$  \\
$\lambda=0, \eta=0.030$ & $854.0$ & $3000.0$ & $363.0$ & $13490.0$  \\
$\lambda=\gamma, \eta=0.005$ & $323.0$ & $3200.0$ & $407.0$ & $0.0$  \\
$\lambda=\gamma,\eta=0.010$ & $421.0$ & $700.0$ & $182.0$ & $13835.0 $ \\
$\lambda=\gamma,\eta=0.015$ & $650.0$ & $3500.0$ & $416.0$ & $14275.0$  \\
$\lambda=\gamma,\eta=0.020$ & $285.0$ & $\mathbf{6800.0}$ & $383.0$ & $13970.0$  \\
$\lambda=\gamma,\eta=0.025$ & $\mathbf{777.0}$ & $\boxed{8700.0}$ & $414.0$ & $13630.0$  \\
$\lambda=\gamma,\eta=0.030$ & $567.0$ & $2450.0$ & $392.0$ & $19745.0$  \\
 \hline
\end{tabular}}
\caption{Final performance for different levels of regularization. Note that the A3C Deepmind scores use a nonpublic human starts evaluation and may not be directly comparable to our random start initialization.}
\end{table}

\begin{figure}[ht!]
  \centering
  \includegraphics[width=\textwidth]{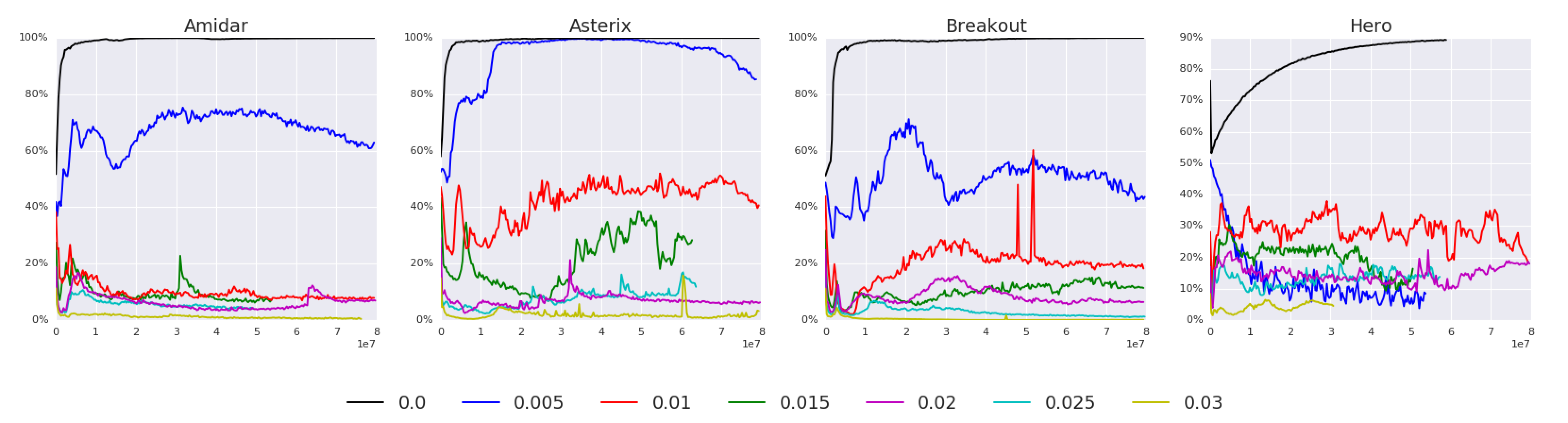}
  \caption{Average termination probabilities through training, with varying amounts of deliberation costs. With no deliberation, the termination rate quickly goes to 100\% (black curve).}
  \label{fig:termplot}
\end{figure}

We also trained agents with multiple levels of deliberation cost, from $\eta=0$ to $\eta=0.03$, with increments of $0.005$. The range of values was chosen according to the general scale of the values proper to these environments. As figure \ref{fig:termplot} shows, an increase in deliberation cost $\eta$ quickly decreases the average termination probabilities as expected by the formulation \eqref{eq:optim}. When no deliberation cost is used, termination raises up to $100\%$ very quickly, meaning each option only lasts a single time-step. The decrease in probability is not the same in every environment, this is due to the difference in returns. The deliberation cost has an effect proportional to its ratio with the state values. Intuitively, environments with many high rewards would indeed require a larger deliberation cost to have substantial effects.

\section{Conclusion and Future Work}

We presented the use of deliberation cost as a way to incentivize the creation of options which persist for a longer period of time. Using this approach in the option-critic architecture yields both good performance as well as options which are intuitive and do not shrink over time. In doing so, we also outlined a connection from our more general notion of deliberation cost with previous notions of regularization from \cite{Mann2014} and \cite{Bacon2017}.

The deliberation cost goes beyond only the idea of penalizing for lengthy computation. It can also be used to incorporate other forms of bounds intrinsic to an agent in its environment. One interesting direction for future work is to also think of deliberation cost in terms of \textit{missed opportunity} and opening the way for an implicit form of regularization when interacting asynchronously with an environment. Another interesting form of limitation inherent to reinforcement learning agents has to do with their representational capacities when estimating action values. Preliminary work seems to indicate that the error decomposition for the action values could be also be expressed in the form of a deliberation cost.

\bibliographystyle{named}
\bibliography{references}

\begin{thebibliography}{}

\bibitem[\protect\citeauthoryear{Altman}{1999}]{Altman1999}
E.~Altman.
\newblock {\em Constrained Markov Decision Processes}.
\newblock Chapman and Hall, 1999.

\bibitem[\protect\citeauthoryear{Andreas \bgroup \em et al.\egroup
  }{2017}]{Andreas2017}
Jacob Andreas, Dan Klein, and Sergey Levine.
\newblock Modular multitask reinforcement learning with policy sketches.
\newblock In {\em ICML}, pages 166--175, 2017.

\bibitem[\protect\citeauthoryear{Bacon \bgroup \em et al.\egroup
  }{2017}]{Bacon2017}
Pierre{-}Luc Bacon, Jean Harb, and Doina Precup.
\newblock The option-critic architecture.
\newblock In {\em AAAI}, pages 1726--1734, 2017.

\bibitem[\protect\citeauthoryear{Baird}{1993}]{Baird1993}
Leemon~C. Baird.
\newblock Advantage updating.
\newblock Technical Report WL--TR-93-1146, Wright Laboratory, 1993.

\bibitem[\protect\citeauthoryear{{Bellemare} \bgroup \em et al.\egroup
  }{2013}]{Bellemare2013}
M.~G. {Bellemare}, Y.~{Naddaf}, J.~{Veness}, and M.~{Bowling}.
\newblock The arcade learning environment: An evaluation platform for general
  agents.
\newblock {\em Journal of Artificial Intelligence Research}, 47:253--279, 06
  2013.

\bibitem[\protect\citeauthoryear{Botvinick \bgroup \em et al.\egroup
  }{2009}]{Botvinick2009}
Matthew~M. Botvinick, Yael Niv, and Andrew~C. Barto.
\newblock Hierarchically organized behavior and its neural foundations: A
  reinforcement learning perspective.
\newblock {\em Cognition}, 113(3):262 -- 280, 2009.

\bibitem[\protect\citeauthoryear{Branavan \bgroup \em et al.\egroup
  }{2012}]{Branavan2012}
S.~R.~K. Branavan, Nate Kushman, Tao Lei, and Regina Barzilay.
\newblock Learning high-level planning from text.
\newblock In {\em ACL}, pages 126--135, 2012.

\bibitem[\protect\citeauthoryear{Daniel \bgroup \em et al.\egroup
  }{2016}]{Daniel2016}
C.~Daniel, H.~van Hoof, J.~Peters, and G.~Neumann.
\newblock Probabilistic inference for determining options in reinforcement
  learning.
\newblock {\em Machine Learning, Special Issue}, 104(2):337--357, 2016.

\bibitem[\protect\citeauthoryear{Dayan and Hinton}{1992}]{Dayan1992}
Peter Dayan and Geoffrey~E. Hinton.
\newblock Feudal reinforcement learning.
\newblock In {\em NIPS}, pages 271--278, 1992.

\bibitem[\protect\citeauthoryear{Dietterich}{1998}]{Dietterich1998}
Thomas~G. Dietterich.
\newblock The {MAXQ} method for hierarchical reinforcement learning.
\newblock In {\em ICML}, pages 118--126, 1998.

\bibitem[\protect\citeauthoryear{Drescher}{1991}]{Drescher1991}
Gary~L. Drescher.
\newblock {\em Made-up Minds: A Constructivist Approach to Artificial
  Intelligence}.
\newblock MIT Press, Cambridge, MA, USA, 1991.

\bibitem[\protect\citeauthoryear{Fikes \bgroup \em et al.\egroup
  }{1972}]{Fikes1972}
Richard Fikes, Peter~E. Hart, and Nils~J. Nilsson.
\newblock Learning and executing generalized robot plans.
\newblock {\em Artif. Intell.}, 3(1-3):251--288, 1972.

\bibitem[\protect\citeauthoryear{Gigerenzer and Selten}{2001}]{Gigerenzer2001}
Gerd Gigerenzer and R.~Selten.
\newblock {\em Bounded Rationality: The adaptive toolbox}.
\newblock Cambridge: The MIT Press, 2001.

\bibitem[\protect\citeauthoryear{Guo \bgroup \em et al.\egroup
  }{2014}]{Guo2014}
Xiaoxiao Guo, Satinder Singh, Honglak Lee, Richard~L Lewis, and Xiaoshi Wang.
\newblock Deep learning for real-time atari game play using offline monte-carlo
  tree search planning.
\newblock In Z.~Ghahramani, M.~Welling, C.~Cortes, N.~D. Lawrence, and K.~Q.
  Weinberger, editors, {\em Advances in Neural Information Processing Systems
  27}, pages 3338--3346. Curran Associates, Inc., 2014.

\bibitem[\protect\citeauthoryear{Howard}{1963}]{Howard1963}
Ronald~A. Howard.
\newblock Semi-markovian decision processes.
\newblock In {\em Proceedings 34th Session International Statistical
  Institute}, pages 625--652, 1963.

\bibitem[\protect\citeauthoryear{Iba}{1989}]{Iba1989}
Glenn~A. Iba.
\newblock A heuristic approach to the discovery of macro-operators.
\newblock {\em Machine Learning}, 3:285--317, 1989.

\bibitem[\protect\citeauthoryear{Jiang \bgroup \em et al.\egroup
  }{2015}]{Jiang2015}
Nan Jiang, Alex Kulesza, Satinder Singh, and Richard~L. Lewis.
\newblock The dependence of effective planning horizon on model accuracy.
\newblock In {\em Proceedings of the 2015 International Conference on
  Autonomous Agents and Multiagent Systems, {AAMAS} 2015, Istanbul, Turkey, May
  4-8, 2015}, pages 1181--1189, 2015.

\bibitem[\protect\citeauthoryear{Kaelbling}{1993}]{Kaelbling1993}
Leslie~Pack Kaelbling.
\newblock Hierarchical learning in stochastic domains: Preliminary results.
\newblock In {\em ICML}, pages 167--173, 1993.

\bibitem[\protect\citeauthoryear{Korf}{1983}]{Korf1983}
Richard~Earl Korf.
\newblock {\em Learning to Solve Problems by Searching for Macro-operators}.
\newblock PhD thesis, Carnegie Mellon University, Pittsburgh, PA, USA, 1983.

\bibitem[\protect\citeauthoryear{Kuipers}{1979}]{Kuipers1979}
Benjamin Kuipers.
\newblock Commonsense knowledge of space: Learning from experience.
\newblock In {\em Proceedings of the 6th International Joint Conference on
  Artificial Intelligence - Volume 1}, IJCAI'79, pages 499--501, San Francisco,
  CA, USA, 1979. Morgan Kaufmann Publishers Inc.

\bibitem[\protect\citeauthoryear{Kulkarni \bgroup \em et al.\egroup
  }{2016}]{Kulkarni2016}
Tejas Kulkarni, Karthik Narasimhan, Ardavan Saeedi, and Joshua Tenenbaum.
\newblock Hierarchical deep reinforcement learning: Integrating temporal
  abstraction and intrinsic motivation.
\newblock In {\em Advances in Neural Information Processing Systems 29}, 2016.

\bibitem[\protect\citeauthoryear{Machado \bgroup \em et al.\egroup
  }{2017}]{Machado2017}
Marlos~C. Machado, Marc~G. Bellemare, and Michael~H. Bowling.
\newblock A laplacian framework for option discovery in reinforcement learning.
\newblock In {\em ICML}, pages 2295--2304, 2017.

\bibitem[\protect\citeauthoryear{Mankowitz \bgroup \em et al.\egroup
  }{2016}]{Mankowitz2016}
Daniel~J. Mankowitz, Timothy~Arthur Mann, and Shie Mannor.
\newblock Adaptive skills, adaptive partitions {(ASAP)}.
\newblock In {\em Advances in Neural Information Processing Systems 29}, 2016.

\bibitem[\protect\citeauthoryear{Mann \bgroup \em et al.\egroup
  }{2014}]{Mann2014}
Timothy~Arthur Mann, Daniel~J. Mankowitz, and Shie Mannor.
\newblock Time-regularized interrupting options {(TRIO)}.
\newblock In {\em ICML}, pages 1350--1358, 2014.

\bibitem[\protect\citeauthoryear{Mann \bgroup \em et al.\egroup
  }{2015}]{Mann2015}
Timothy~Arthur Mann, Shie Mannor, and Doina Precup.
\newblock Approximate value iteration with temporally extended actions.
\newblock {\em J. Artif. Intell. Res.}, 53:375--438, 2015.

\bibitem[\protect\citeauthoryear{Minsky}{1961}]{Minsky1961}
Marvin Minsky.
\newblock Steps toward artificial intelligence.
\newblock {\em Proceedings of the IRE}, 49(1):8--30, January 1961.

\bibitem[\protect\citeauthoryear{Mnih \bgroup \em et al.\egroup
  }{2015}]{mnih2015human}
Volodymyr Mnih, Koray Kavukcuoglu, David Silver, Andrei~A Rusu, Joel Veness,
  Marc~G Bellemare, Alex Graves, Martin Riedmiller, Andreas~K Fidjeland, Georg
  Ostrovski, et~al.
\newblock Human-level control through deep reinforcement learning.
\newblock {\em Nature}, 518(7540):529--533, 2015.

\bibitem[\protect\citeauthoryear{Mnih \bgroup \em et al.\egroup
  }{2016}]{mnih2016asynchronous}
Volodymyr Mnih, Adria~Puigdomenech Badia, Mehdi Mirza, Alex Graves, Timothy
  Lillicrap, Tim Harley, David Silver, and Koray Kavukcuoglu.
\newblock Asynchronous methods for deep reinforcement learning.
\newblock In {\em ICML}, pages 1928--1937, 2016.

\bibitem[\protect\citeauthoryear{Neyman}{1985}]{Neyman1985}
Abraham Neyman.
\newblock Bounded complexity justifies cooperation in the finitely repeated
  prisoners dilemma.
\newblock {\em Economics Letters}, 19(3):227--229, jan 1985.

\bibitem[\protect\citeauthoryear{Parr and Russell}{1998}]{Parr1997}
Ronald Parr and Stuart~J. Russell.
\newblock Reinforcement learning with hierarchies of machines.
\newblock In M.~I. Jordan, M.~J. Kearns, and S.~A. Solla, editors, {\em
  Advances in Neural Information Processing Systems 10}, pages 1043--1049. MIT
  Press, 1998.

\bibitem[\protect\citeauthoryear{Petrik and Scherrer}{2008}]{Petrik2008}
Marek Petrik and Bruno Scherrer.
\newblock Biasing approximate dynamic programming with a lower discount factor.
\newblock In {\em Advances in Neural Information Processing Systems 21,
  Proceedings of the Twenty-Second Annual Conference on Neural Information
  Processing Systems, Vancouver, British Columbia, Canada, December 8-11,
  2008}, pages 1265--1272, 2008.

\bibitem[\protect\citeauthoryear{Precup}{2000}]{Precup2000}
Doina Precup.
\newblock {\em Temporal abstraction in reinforcement learning}.
\newblock PhD thesis, University of Massachusetts Amhersts, 2000.

\bibitem[\protect\citeauthoryear{Puterman}{1994}]{Puterman1994}
Martin~L. Puterman.
\newblock {\em Markov Decision Processes: Discrete Stochastic Dynamic
  Programming}.
\newblock John Wiley \& Sons, Inc., New York, NY, USA, 1994.

\bibitem[\protect\citeauthoryear{Sennott}{1991}]{Sennott1991}
Linn~I. Sennott.
\newblock Constrained discounted markov decision chains.
\newblock {\em Probability in the Engineering and Informational Sciences},
  5(4):463–475, 1991.

\bibitem[\protect\citeauthoryear{Simon}{1957}]{Simon1957}
Herbert~A. Simon.
\newblock {\em Models of man: social and rational; mathematical essays on
  rational human behavior in society setting}.
\newblock Wiley, 1957.

\bibitem[\protect\citeauthoryear{Solway \bgroup \em et al.\egroup
  }{2014}]{Solway2014}
Alec Solway, Carlos Diuk, Natalia Córdova, Debbie Yee, Andrew~G. Barto, Yael
  Niv, and Matthew~M. Botvinick.
\newblock Optimal behavioral hierarchy.
\newblock {\em PLOS Computational Biology}, 10(8):1--10, 08 2014.

\bibitem[\protect\citeauthoryear{Sutton \bgroup \em et al.\egroup
  }{1999a}]{Sutton1999}
Richard~S. Sutton, David~A. McAllester, Satinder~P. Singh, and Yishay Mansour.
\newblock Policy gradient methods for reinforcement learning with function
  approximation.
\newblock In {\em NIPS}, pages 1057--1063, 1999.

\bibitem[\protect\citeauthoryear{Sutton \bgroup \em et al.\egroup
  }{1999b}]{SuttonPrecupSingh1999}
Richard~S. Sutton, Doina Precup, and Satinder~P. Singh.
\newblock Between mdps and semi-mdps: {A} framework for temporal abstraction in
  reinforcement learning.
\newblock {\em Artif. Intell.}, 112(1-2):181--211, 1999.

\bibitem[\protect\citeauthoryear{Sutton}{1984}]{Sutton1984}
Richard~S. Sutton.
\newblock {\em Temporal credit assignment in reinforcement learning}.
\newblock PhD thesis, University of Massachusetts Amherst, 1984.

\bibitem[\protect\citeauthoryear{Thrun and Schwartz}{1995}]{Thrun95}
Sebastian Thrun and Anton Schwartz.
\newblock Finding structure in reinforcement learning.
\newblock In {\em NIPS}, 1995.

\end{thebibliography}

\clearpage
\onecolumn
\section{Appendix}

\begin{figure}[ht!]
    \centering
  \includegraphics[width=\textwidth]{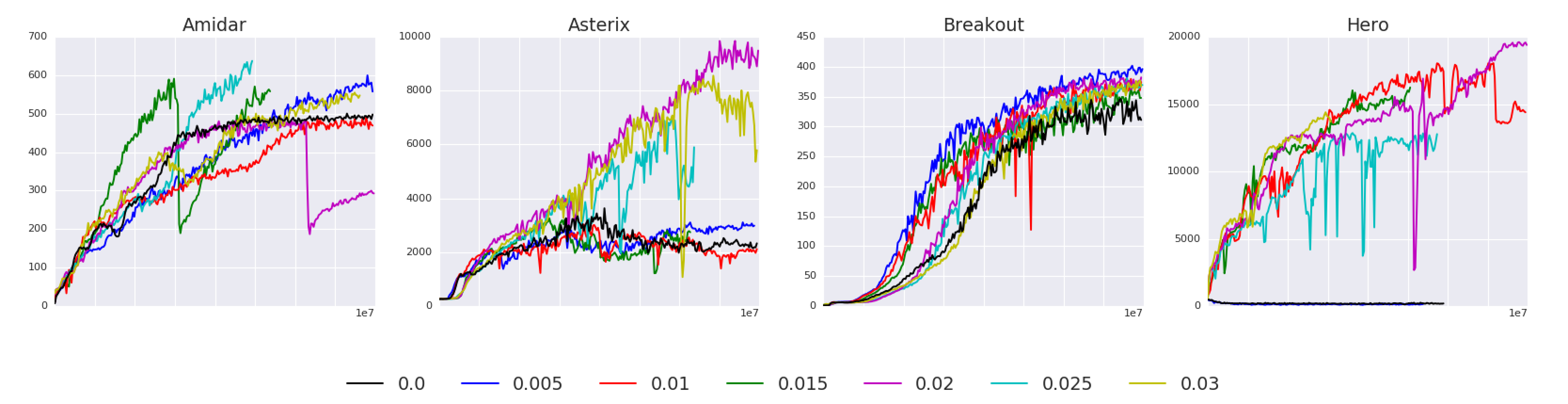}
  \caption{Training curves with different deliberation costs on 4 Atari 2600 games. Trained for up to 80M frames.}
\end{figure}

\end{document}